\documentclass[runningheads]{llncs}
\usepackage{graphicx}
\usepackage[style=numeric,natbib=true]{biblatex}
\usepackage{booktabs}
\usepackage{tabularx}
\usepackage{amsmath}
\usepackage{adjustbox}
\usepackage{todonotes}
\usepackage{hyperref}
\usepackage{cleveref}
\usepackage{subcaption} 
\usepackage{eurosym}
\usepackage[normalem]{ulem}
\useunder{\uline}{\ul}{}
\newcommand*\rot{\rotatebox{90}}
\begin{document}
\title{Open ERP System Data For Occupational Fraud Detection}
%
%
\author{Julian Tritscher\inst{1} \and
Fabian Gwinner\inst{2} \and
Daniel Schlör\inst{1} \and
Anna Krause\inst{1} \and
Andreas Hotho\inst{1}}
\authorrunning{J. Tritscher et al.}
%
\institute{University of Würzburg, Am Hubland, 97074 Würzburg, Germany\\
\email{\{tritscher, schloer, anna.krause, hotho\}@informatik.uni-wuerzburg.de} \and 
\email{fabian.gwinner@uni-wuerzburg.de}}
\maketitle              
\begin{abstract}
Recent estimates report that companies lose $5\%$ of their revenue to occupational fraud.
Since most medium-sized and large companies employ Enterprise Resource Planning (ERP) systems to track vast amounts of information regarding their business process, researchers have in the past shown interest in automatically detecting fraud through ERP system data.
Current research in this area, however, is hindered by the fact that ERP system data is not publicly available for the development and comparison of fraud detection methods.
We therefore endeavour to generate public ERP system data that includes both normal business operation and fraud. 
We propose a strategy for generating ERP system data through a serious game, model a variety of fraud scenarios in cooperation with auditing experts, and generate data from a simulated make-to-stock production company with multiple research participants. 
We aggregate the generated data into ready to used datasets for fraud detection in ERP systems, and supply both the raw and aggregated data to the general public to allow for open development and comparison of fraud detection approaches on ERP system data.
\end{abstract}

\section{Introduction}
The association of certified fraud examiners defines occupational fraud as abusing one's occupation through the deliberate abuse of an employing organization's assets, and estimates that currently companies lose 5\% of their revenue to this type of fraud \citep{ACFE2022}.
To reduce the loss of revenue to occupational fraud, researchers have in the past suggested to use the data contained within ERP systems to detect fraudulent activity \citep{singh2016interactive,schreyer2017erp,Oliverio2019Oct}.
ERP systems are a core component for managing the flows of cash, materials, production and other resources within companies. 
They represent a large market with $37,679.26$ Mio. USD worldwide revenue in 2020, and support most medium-sized and large companies in their daily work \citep{statisticaloffice2020ICT,Statista2020revenue}. 

In spite of ERP systems providing many different views on an organization's workflow that could potentially aid the detection of fraudulent activity, research in this area is currently hindered by the fact that ERP system data, in general, is not available to the public. 
This proves problematic when attempting to reproduce published results, and compare performance of existing fraud detection approaches.

To address this issue, we propose an approach for generating synthetic ERP system data that extends previous research, generating data containing both normal operation and fraudulent activities, and making the resulting data publicly available.

Previous works on ERP system fraud detection may be divided into approaches that rely on entirely private data and frauds, private data with synthetically injected frauds, or entirely synthetic data and frauds.
While there have been works that use real ERP system data to develop and evaluate fraud detection systems \citep{singh2016interactive,schreyer2017erp,Oliverio2019Oct}, details about the data and the data itself are kept under wraps to avoid revealing company trade secrets and privacy information. 
For scenarios where real frauds are not available, \citet{islam2010ERPfraud} generate synthetic fraud cases within private ERP system data through randomly creating changes to normal transactions while limiting changes and timings with given intervals. 
While this generates anomalous transactions through not yet observed peaks in single entries, the generated anomalies have no inherent meaning or interpretation with respect to real-life occurrences or fraud. 

As an alternative that uses no private data, business researchers have in the past moved to developing data generators that are capable of generating both normal operation and frauds: 
\citet{yannikos20103lspg} introduce 3LSPG, a generator that produces synthetic ERP data through a probabilistic approach using discrete time Markov chains. 
While the resulting data can mimic the transactions taken from an ERP system, the data's quality and realism are strongly dependent on the expert knowledge put into the simulation.
With no data, code, and chosen simulation parameters available, modeling realistic ERP system data through this approach is challenging.
Similarly, game based approaches may be used to model business processes with player interaction \citep{tritscher2021fraudGame}. 
While this is a promising aspect for data generation, expert knowledge is required to ensure that the complex behavior of real ERP systems are mimicked in the resulting data.


%
\citet{baader2016dataGenerator} partially alleviate the need of expert knowledge by modeling normal and fraudulent behavior directly within an ERP system, thus being able to automate parts of the generation process that would be carried out by the ERP system in a real world scenario. Remaining business decisions that are not taken over by the ERP system such as procurement quantities or sales prices are simulated through random distributions.
\citet{baader2018syntheticERPdata} extend the approach in additional work, where, instead of generating fraud cases through random distributions, they obtain fraud scenarios through user participation with the white-collar hacking contest \citep{schermann2014hackingContest}, a serious game developed to teach players the abuse of an ERP system and the detection thereof.
While the resulting frauds may model realistic scenarios, they are modeled into an existing database in post, potentially causing unwanted divergence between normal and fraudulent data characteristics.
Further, in contrast to other research areas that utilise synthetic data such as intrusion detection \citep{ring2019survey}, all published synthetic ERP data generation methods to our knowledge do not publish their code and data, making reproducible and incremental research in ERP fraud detection difficult due to missing comparability.

In this paper, we address these problems by extending the work of \citet{baader2018syntheticERPdata} in multiple ways:
We first extend the requirements for synthetic ERP data layed out by \citet{baader2016dataGenerator} to include further requirements of ERP fraud detection approaches \citep{fuchs2021ERPfraud}.
We secondly propose to use an established serious game \citep{leger2007erpsim} to simulate not only fraudulent scenarios but also normal operation of a make-to-stock production company through user interaction in a real ERP system, generating both normal and fraudulent behavior simultaneously. 
Additionally, this allows us to extend the business processes investigated by previous data generators from the purchase-to-pay (P2P) process to modeling normal and fraudulent activity in the well established order-to-cash (O2C) process as well.
Based on our extended requirements, we then design multiple fraud scenarios in cooperation with auditing experts.
We conduct multiple runs of our proposed data generation scheme and produce ERP system data of multiple fiscal years of operation, extracting raw data from the ERP system.
The resulting data contains many multi-relational tables that offer different views on the recorded company's business process.
Since many fraud detection approaches require single tables to operate, we additionally create ready to use datasets from a subset of our multi-relational data that can be directly used for measuring and comparing the performance of fraud detection systems.
We further extend these datasets by providing expert-created annotations for fraud cases that highlight the problematic entries of individual frauds for use in debugging and assessing performance of algorithms that focus on the detection of anomalous entries specifically.

In summary, our main contributions are as follows:
\begin{itemize}
    \item We propose a strategy for data generation that simulates normal behavior and fraud jointly through user interaction within a real ERP system and is capable of modeling frauds and normal behavior in the P2P and O2C business processes.
    \item We conduct multiple simulation runs and construct ready to use datasets with detailed fraud annotations that allow for direct application and comparison of ERP fraud detection approaches. 
    \item Finally, we provide both raw data and ready to use datasets to the general public to allow for open comparability of ERP fraud detection systems.\footnote{Datasets are available under \url{https://professor-x.de/erp-fraud-data}.}
\end{itemize}

The remaining paper is structured as follows: 
\Cref{chap:data} outlines the requirements for data generation, introduces our data generator, showcases the modelled business scenario and chosen fraud cases, and details the data generation process.
\Cref{chap:analysis} presents an analysis on the collected data, while \Cref{chap:dataset_construction} aggregates parts of the data into ready to use datasets for direct use in fraud detection applications.
\Cref{chap:conclusion} concludes the paper.

\section{Data Collection}
\label{chap:data}

\begin{figure}[t]
    \centering
    \includegraphics[width=1\textwidth]{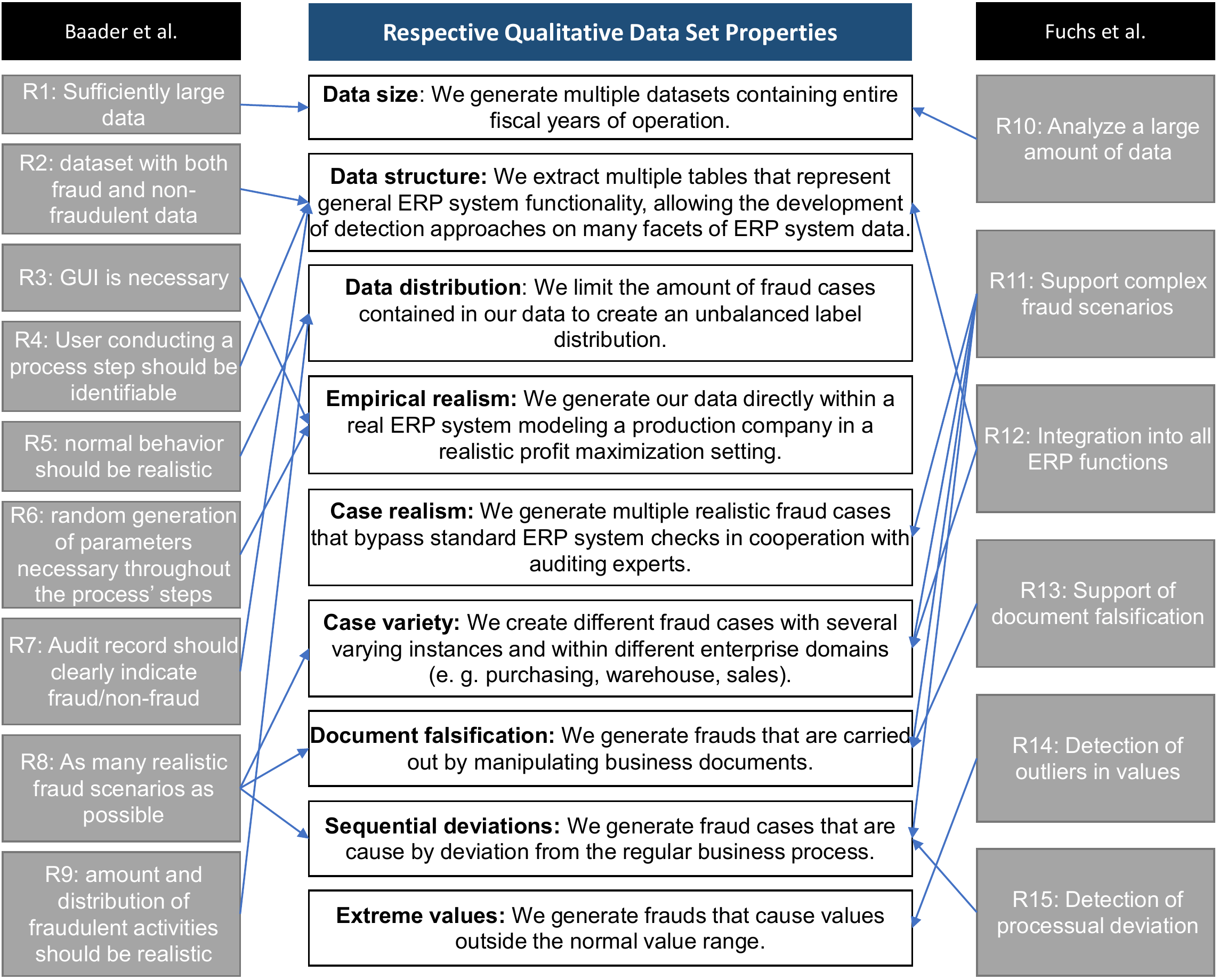}
    \caption{Dataset properties derived from literature requirements.} 
    \label{fig:requirements}
\end{figure}

\subsection{Data Requirements}
\label{data:requirements}
To create high-quality ERP system data and increase rigor, we develop requirements for our data based on prior work. We follow \citet{baader2016dataGenerator}, who find several requirements for both their developed data generator and the resulting data. We also draw additional requirements from previously conducted design science research that aims to develop a fraud detection system in the ERP domain. Here, \citet{fuchs2021ERPfraud} aggregate requirements for detection systems that are able to highlight fraud in ERP systems.    
Some of their requirements describe design decisions that need to be respected during implementation of the fraud detection approach and are unaffected by the studied data (e.g. requiring adaptable or intelligent logic). 
Other requirements, however, describe scenarios in which fraud detection approaches should yield satisfying performance (e.g. detection of outliers in values).  
We argue that data should be created such that the performance of fraud detection approaches can be validated in these scenarios, and therefore identify these requirements as directly relevant for our data generation process.

\Cref{fig:requirements} gives an overview of the requirements in the preliminary work of \citet{baader2016dataGenerator} as well as the requirements we identify as relevant to our data generation process from \citet{fuchs2021ERPfraud}. 
We additionally note the resulting measures we take in our proposed data generation scheme to satisfy these requirements.

\subsection{Data Generation through ERPsim}
\label{data:erpsim}
Similar to \citet{baader2018syntheticERPdata} that use an existing serious game \citep{schermann2014hackingContest} to generate fraudulent ERP transactions through user interaction, we take a game-based data generation approach to meet our formal data generation requirements and employ a serious game, ERPsim \citep{leger2007erpsim}, to record ERP system data.
Within the ERPsim serious game, participants take control of a make-to-stock production company through the ability to plan overall sales, create purchase orders for raw materials, plan production of products, produce and deliver sales orders to fictitious customers, manage the accounting, and optionally take loans and manage debts. 
In our scenario, the ERP system is used for make-to-stock production of four products based on a market analysis and forecast. After production, products are stored in a warehouse and sold to customers. 
To simulate an in-game year for a single company, the game may be played with up to five players per company in the roles of material planner, production controller, sales manager, financial planner, and market analyst.

We choose ERPsim, as it allows for generating both normal operation and fraudulent activities through user interaction.
Next to the added complexity that may be introduced to the data through multiple participants operating the company simultaneously, ERPsim also offers a realistic profit maximization scenario through a simulated market.
Where business decisions such as deciding on purchase quantities are modeled through random distributions in prior data generators \citep{baader2016dataGenerator}, ERPsim motivates participants to make economically sensible business decisions.
The game is also conducted directly within an SAP S/4 HANA ERP system. 
While real life processes such as the delivery of raw materials and the production of goods are simulated, the resulting documents and transactions are recorded within the ERP system through the standardized processes that are also employed in real companies.
Unlike previous work, this allows fraud scenarios to be committed directly in the ERP system interface during operation, rather than requiring anomalies to be synthetically injected into historical data of normal operation.

\subsection{Modeled Business Scenario}
\label{data:scenario}
As determined through our requirements analysis in \Cref{data:requirements}, data for ERP fraud detection requires a large variety of realistic fraud cases. 
To create realistic fraud cases that translate well to different companies, we focus on creating frauds within two standardized business processes that are simulated within our data generator. 

\begin{figure}[t]
    \centering
    \includegraphics[width=1\textwidth]{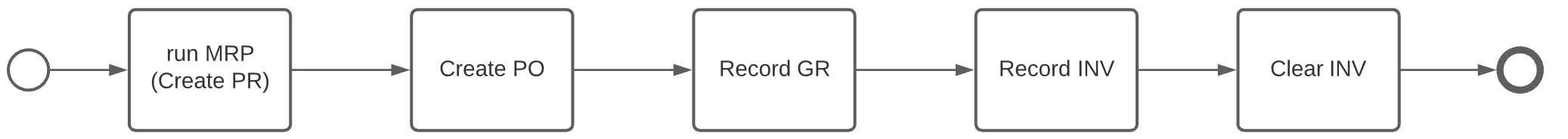}
    \caption{Purchase-to-pay (P2P) process in the ERPsim simulation.} 
    \label{fig:p2p_steps}
\end{figure}

First, we select the widely used purchase-to-pay (P2P) process that has been the focus of previous data generation approaches \citep{yannikos20103lspg,baader2016dataGenerator,baader2018syntheticERPdata}. 
In the P2P scenario, illustrated in \Cref{fig:p2p_steps}, a demand is created by the user's forecast. 
By performing the Material Requirement Planning (MRP) run, the user creates a purchase requisition (PR) for each demand. 
A buying agent then converts each PR into a purchase order (PO). 
As in the real world, saving a PO starts transferring the PO to the given supplier, where it gets converted into a customer order. 
While these steps are usually implemented via electronic data interchange between two ERP systems, in ERPsim the simulation middle-ware receives the PO and virtually ships the ordered goods after a random time within a defined time-frame. 
After this, the incoming goods need to be received at the production plant with a goods receipt (GR), and paid for by recording and clearing the invoice (INV).

\begin{figure}[b]
    \centering
    \includegraphics[width=1\textwidth]{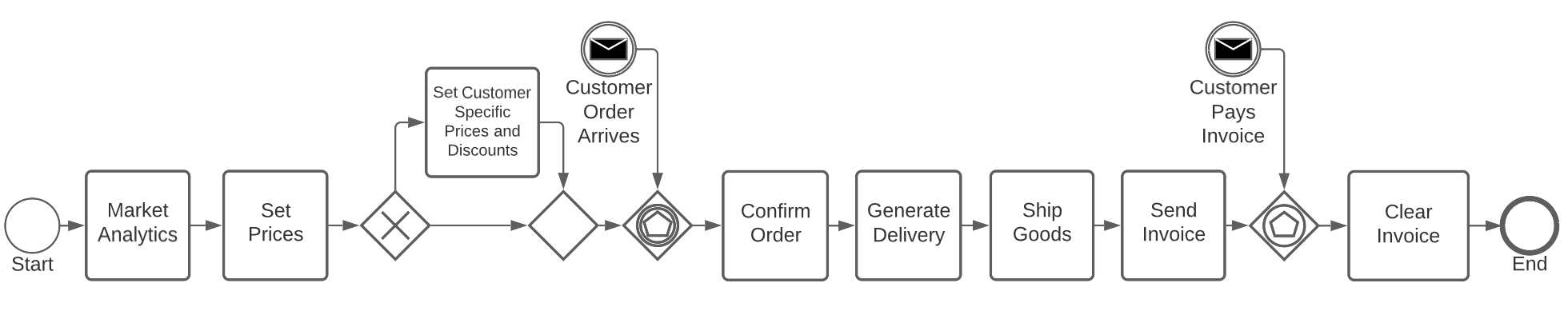}
    \caption{Order-to-cash (O2C) process in the ERPsim simulation.} 
    \label{fig:o2c_steps}
\end{figure}

As second business process, we select the well established order-to-cash (O2C) process that allows for modelling frauds in the sales department and has been suggested as future work for simulation by \citet{baader2016dataGenerator}. 
The O2C process is usually comprised of the activities from the customer ordering products to the payment of the order. 
As prices are an important component in our modeled fraud cases, we add the activities to determine sales prices into our O2C process. 
The resulting O2C process is illustrated in \Cref{fig:o2c_steps} and consists of the market analysis and price calculation for determining the overall prices of sales products. 
Afterwards, customer or market specific discounts can be determined. 
Based on the resulting sales prices, a customer demand may be generated in the market and an order is generated. 
The orders then can be viewed, altered and confirmed. 
Confirming the order then triggers an activity for creating a delivery that ships the goods to the customer and generates an invoice. 
After a randomized time the customer pays the invoice and the accountant clears the open invoice, ending the process.

\begin{figure}[b]
    \centering
    \includegraphics[width=1\textwidth]{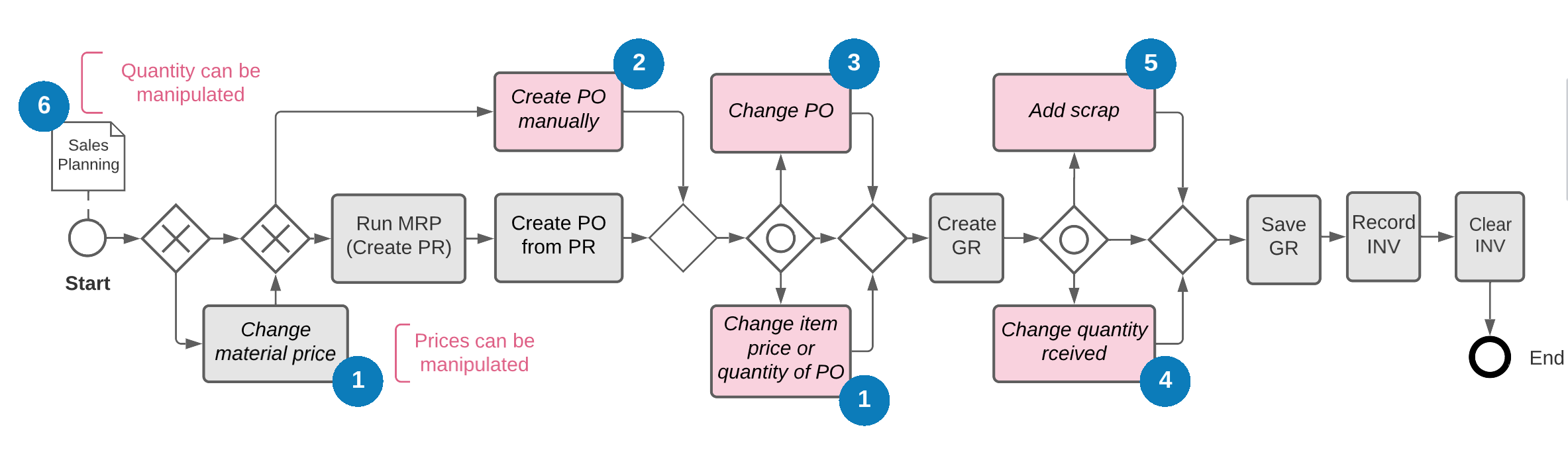}
    \caption{P2P process steps with potential fraudulent deviations.} 
    \label{fig:p2p_frauds}
\end{figure}

\subsection{Modeled Fraud Cases}
\label{data:frauds}
With the standardized business processes selected and introduced, we now turn to modeling fraudulent activity within these processes.
Next to the requirement of a large variety of frauds, our analysis of \citet{fuchs2021ERPfraud} yields specific types of deviation that need to be included in the data.
To satisfy the data requirements regarding fraud cases, we first analyse the fraud scenarios conducted by \citet{baader2016dataGenerator}. 
While analysing their modeled fraud scenarios, we found multiple cases are not possible in our realistic environment, as our ERP system is equipped with realistic control and audit mechanisms (e.g. due diligence, user authorisations, accounting checks), that are integrated in most ERP systems and are employed in many companies due to best practices or legal regulations \citep{osaci2018sap,mans2008application,lightle2003segregation,act2002sarbanes}. 
Since these control mechanisms prevent some cases of undesirable user behaviour, many fraud cases from \citet{baader2016dataGenerator} such as double payment, conto pro diverse transaction fraud, or non-purchase payment, would be blocked by the ERP system and would at most yield unsuccessful fraud attempts. 
Some further fraud cases of \citet{baader2016dataGenerator} rely on abusing quantity contracts for suppliers or value contracts for customers. 
As in our scenario purchase prices are driven by a simulated market where prices for raw materials change frequently, our use case does not provide any framework for contracts.
Therefore, frauds involving contracts could not be simulated. 
The remaining fraud cases "false invoice fraud" (in our scenario Larceny 1 and 2) and "misappropriation fraud" (Larceny 4) are part of our modeled fraud cases. 

After the preliminary selection of three fraud cases from \citet{baader2016dataGenerator}, we select nine additional fraud scenarios to match our identified requirements.
Here we select appropriate fraud scenarios from the ACFE's report to the nations \citep{ACFE2022} in cooperation with business auditing experts.
Since, in contrast to \citet{baader2016dataGenerator}, our data generation approach is also capable of modeling the entire O2C business process, we additionally take care to include fraud scenarios that are conducted within the O2C process.

In total, we obtain twelve fraud cases that represent a broad spectrum of fraud on data level.
Eight of the selected twelve fraud cases are part of the P2P process.
In \Cref{fig:p2p_frauds} we visualize the fraudulent deviations (red activities) that may be used by these fraud cases to deviate from the normal business process (grey activities).

\begin{figure}[t]
    \centering
    \includegraphics[width=1\textwidth]{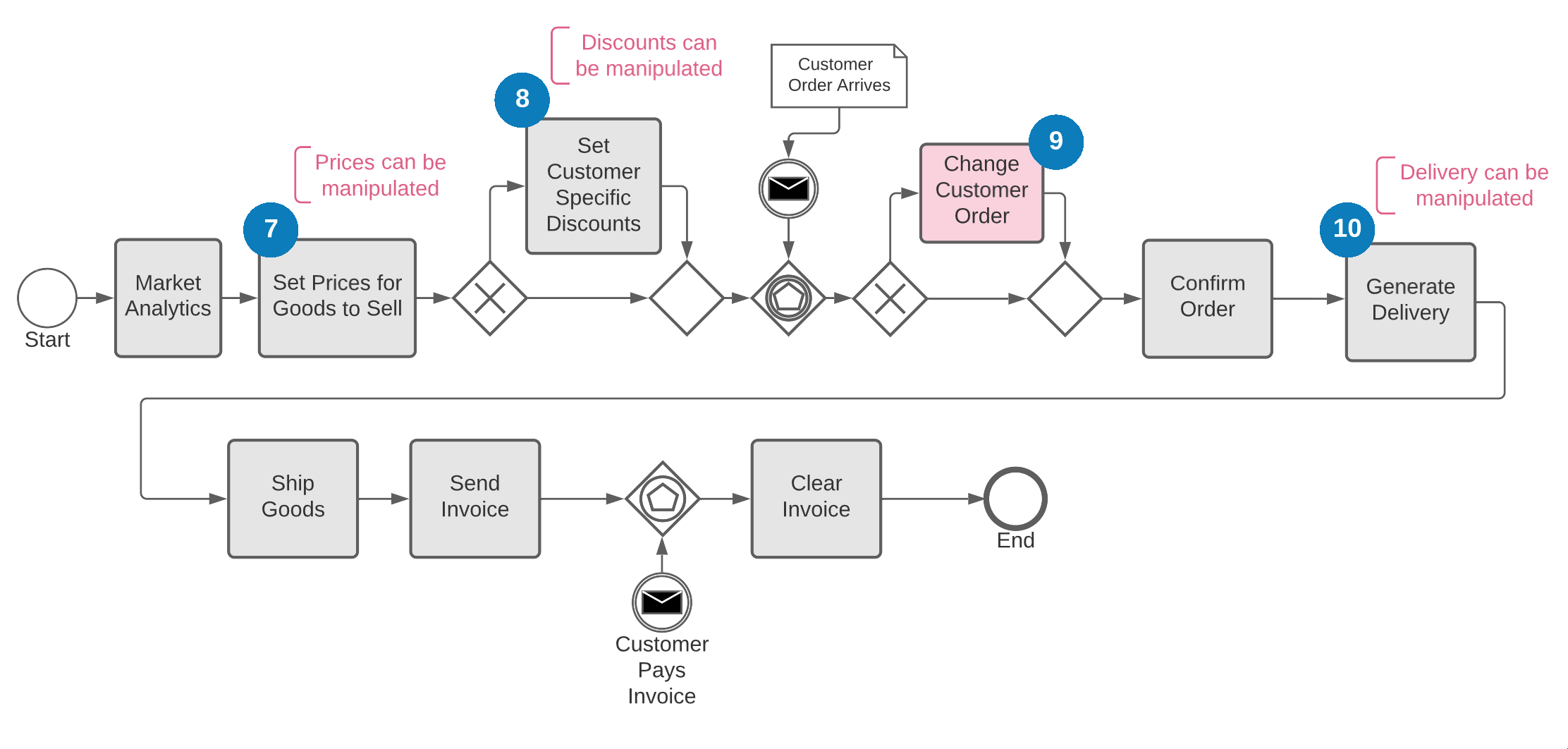}
    \caption{O2C process steps with potential fraudulent deviations.} 
    \label{fig:O2C_frauds}
\end{figure}

The remaining four fraud cases are part of the O2C process, where we visualize the fraudulent deviations in the same way in \Cref{fig:O2C_frauds}.

An in-depth description of all frauds and their used deviations from the normal process are detailed in Table~\ref{tab:fraud_cases}.
To ease readability, we categorize our different fraud scenarios as follows.
We denote frauds as \emph{invoice kickback} frauds that attempt to falsify invoice documents in order to create an advantage for an accomplice supplier that may be shared with the fraudster.
Frauds that focus on the theft of materials from the company are noted as \emph{larceny} frauds.
\emph{Corporate injury} frauds describe frauds that do not yield monetary benefit for the fraudster, but instead aim to cause financial harm the company.
Finally, \emph{selling kickback} frauds represent fraudsters manipulating sales conditions for accomplice customers to gain potentially shared financial benefits.


\setlength{\tabcolsep}{6pt}
\begin{table}[!htbp]
  \centering 
    \caption{Overview of chosen fraud scenarios with numbered deviations visualized in \Cref{fig:p2p_frauds,fig:O2C_frauds}.}
    \resizebox{0.95 \textwidth}{!}{ 
    \begin{tabularx}{\textwidth}{p{2.7cm} X l}
    \toprule
    Fraud               & Description  \\ 
    \midrule
    Invoice Kickback 1 \newline (P2P)   & Item purchase prices are changed while or before creating a PO (1) for an existing PR, to get a kickback from the supplier later. Results are higher unit prices, causing an anomaly where the combination of quantities and amounts diverge from the normal data distribution.  \\
    Invoice Kickback 2 \newline (P2P)   & Instead of changing an existing PR as in Invoice Kickback 1, an old PO is copied manually (2) with higher prices, resulting in manually created transactions. \\
    Larceny 1  \newline (P2P)           & Quantities of purchased items are changed in an existing PO (3). While recording the GR, the expected item quantity is changed back (4) to leave no leftover in stock. A rule-based ERP check found goods are missing and internally blocked this transaction, making this an unsuccessful fraud attempt. \\
    Larceny 2 \newline (P2P)           & Similar to Larceny 1, but here the PO was released manually (2) and changed afterwards (1), and the GR was booked regularly. The system’s rule-based approach was unsuccessful, leading to an increase in inventory without the items’ physical presence. Although this could be detected during stocktaking, damage would have already been caused to the company. Structurally, this fraud causes anomalously missing values due to manual PO creation. \\
    Larceny 3 \newline (P2P)          & In this case, goods are purchased regularly and a partial amount of waste or scrap is booked (5) through the quality inspection in the goods receipt activity, to hide the theft of goods. This case may also be applied in warehousing or production. \\
    Larceny 4 \newline (P2P)          & Here, goods that are usually not needed for production or organizational processes are purchased through a manual PO (2). While many companies use a four eyes principle (two-man rule), we assume a collusion of purchaser and supervisor. \\
    Larceny 5 \newline (P2P)         & In this fraud, products were ordered regularly, but the delivery address was changed for the PO (3) to a private address. \\ 
    Larceny 6 \newline (O2C)         & Similar to larceny 5, but in the O2C process. The delivery address in the master data of a customer in the order was changed (9), so that the delivery of a corresponding customer order was delivered to the wrong address. The address was changed back afterwards.  \\
    %
    %
    Corporate Injury 1 \newline (P2P)   & This fraud represents extensively large purchases by changing the Sales Planning (6), leading to company damages through high spending and potential waste and overstocking of warehouses. Structurally, this fraud results in anomalous extreme values in POs. \\
    Corporate Injury 2 \newline (O2C)  & Here, the employee committing fraud drastically lowered sales prices (7) to damage the company. \\
    Selling Kickback 1 \newline (O2C)  & This fraud case was conducted by manipulating sales conditions (8) for specific customers, which allowed the customers to purchase products with lowered order prices via discounts. \\
    Selling Kickback 2 \newline (O2C)  & Similar to Selling Kickback 1, the order prices are manipulated. In this case, beneficial sales conditions were given to a specific customer in the sales order document itself (9). \\
    \bottomrule
    \end{tabularx}%
    }
  \label{tab:fraud_cases}%
\end{table}%

\subsection{Data Generation}
\label{data:generation}
Using our proposed data generation process and the collected frauds, we conduct multiple runs of the ERPsim game to generate ERP system data, with each run generating data of one fiscal year of operation.
Runs are played by five research participants with an information systems background.
Participants are instructed on the business process specifics of the company modeled within ERPsim and adopt the roles described in \Cref{data:erpsim}.

To model the proposed frauds during our data generation process, we assign one participant the role of fraudster who may introduce fraudulent activities throughout the data generation process.
The fraudster is instructed by an auditing expert on how to conduct our chosen fraud cases within the ERP system interface.
We further identified in \Cref{data:requirements} that the number and distribution of committed frauds should be realistic.
\citet{schreyer2017erp} argue that real audit scenarios have highly unbalanced class distribution between very few anomalous and vast amounts of regular entries.
Judging the real number of occupational fraud cases that are expected to lie within a company's data, however, is challenging, since the number of employees engaging in fraud is unknown and even in detected frauds the large majority of cases contains active attempts to hide the fraudulent activity \cite{ACFE2022}.
To limit the amount of frauds included within our data and obtain a heavily unbalanced class distribution, we therefore limit our fraudster to conducting two fraud cases per simulated month of operation.

In this setting, we conduct multiple data generation runs in the SAPs R/3 on HANA ERP system together with ERPsim R11.2 with a group of 5 research participants. 
We let the group play the game twice, obtaining a run of exclusively normal operation (normal 1) as well as a run that has fraudulent activities incorporated next to normal business processes (fraud 1).
To obtain differing company characteristics such as varying business strategies and user behavior, we additionally select a second group of participants to generate data.
Our second group of participants generate one run of normal operation (normal 2) and two individual runs containing different fraud cases (fraud 2, 3), resulting in 3 datasets, each simulating one financial year. 

To further increase the complexity of our generated data, we extend the normal business procedure of ERPsim by modeling specific events with our second participant group.
As some of our modeled fraud cases involve process steps that are usually not part of the ERPsim game, such as booking of scrap or giving customer discounts, we add these behaviors as additional activities, build them as repetitive tasks into the process and track them similarly to the conducted fraud cases.
For scrap we add regular manual scrap bookings of received goods, to simulate problems in the delivery or warehouse operations, while limiting the bookings to small amounts of broken materials.
To simulate a regularity in customer discounts we add promotional campaigns, that give customer groups a small discount for their purchases within a given time frame, through setting the appropriate discounts for the distribution channels within the ERP system.

\section{Analysis of Generated Data}
\label{chap:analysis}

Within our proposed data generation scheme, we conducted five separate runs of the ERPsim serious game with both exclusively normal and partially fraudulent business operation.
In \Cref{tab:data_characteristics} we report some financial characteristics of the simulated company over the conducted runs, with each run lasting one fiscal year.
When comparing purchasing costs and turnover costs, we observe that all companies were capable of achieving a considerable added value through the procurement, production, and sales strategies employed by the data generation participants.
Our first participant group was capable of achieving higher added values within their runs normal 1 and fraud 1, which can be attributed to a largely differing business strategy compared to our second group.
While our second group specifically targeted large resellers in their runs (normal 2, fraud 2, fraud 3) through producing exclusively large product sizes and was capable of serving the market of the $71$ large resellers within ERPsim, our first group produces additional small product sizes that are sold also to smaller retails which left them with a higher number of customers.
This also explains the high turnover volume in comparison to the purchasing volume of run normal 1, as turnover volume here also included smaller packaging.
Overall, we find that the different participant groups indeed generated data with varying characteristics through the choice of different business strategies.

\begin{table}[t]
\caption{Data characteristics of the recorded runs of ERPsim $^*$\textit{excluding fraud}}
\label{tab:data_characteristics}
\resizebox{\textwidth}{!}{
\begin{tabular}{lrrrrrrr}
\toprule
Dataset  & 
\begin{tabular}[c]{@{}c@{}}Turnover \\ volume (qty)\end{tabular} & 
\begin{tabular}[c]{@{}c@{}}Turnover \\ costs (\euro)\end{tabular} & 
\begin{tabular}[c]{@{}c@{}}Purchasing \\ volume (kg)\end{tabular} & 
\begin{tabular}[c]{@{}c@{}}Purchasing \\ costs$^*$(\euro)\end{tabular} & 
\begin{tabular}[c]{@{}c@{}}No. of \\ customers\end{tabular} & 
\begin{tabular}[c]{@{}c@{}}No. of \\ sales\end{tabular} & 
\begin{tabular}[c]{@{}c@{}}No. of \\ purchases\end{tabular} \\
\midrule
normal 1 & $3,382,416$    & $19,482,620,84$   & $2,764,010$         & $6,088,646.20$         & $194$   & $9,329$    & $552$     \\
fraud 1  & $2,655,058$    & $13,740,944.11$   & $6,598,400$         & $3,654,801.50$         & $194$   & $6,605$    & $222$     \\ 
normal 2 & $2,925,000$    & $12,901,063.42$   & $2,915,200$         & $6,424,043.80$         & $71$    & $6,243$    & $257$     \\
fraud 2  & $3,026,318$    & $14,154,144.48$   & $3,667,800$         & $7,090,319.80$         & $71$    & $6,793$    & $287$     \\
fraud 3  & $3,244,421$    & $14,820,360.15$   & $4,727,500$         & $7,133,712.35$         & $71$    & $7,113$    & $280$     \\\hline
\end{tabular}}
\end{table}

Beyond the economic characteristics of the simulated companies, we report the number of fraud cases and added additional events within the generated data in \Cref{tab:data_frauds}.
As described in \Cref{data:generation}, the total amount of fraud cases was kept low to retain an unbalanced data distribution.
All fraud runs contain fraud scenarios within the P2P and O2C business process.
Our first participant group focused on several scenarios of larceny fraud.
Our second group, on the other hand, modeled multiple fraud scenarios that hide their activities through scrap bookings, and frauds that achieve profit through fraudulent discounts.
In their runs (normal 2, fraud 2, fraud 3) we also simulated regular scrap bookings and sales events as described in \Cref{data:generation}.

Overall, both groups modeled a variety of complex fraud scenarios within their generated data, with different distributions of fraud cases.

\begin{table}[b]
\caption{Fraud cases and events occurring within the recorded runs of ERPsim.}
\label{tab:data_frauds}
\resizebox{\textwidth}{!}{
\begin{tabular}{lrrrrrrrrrrrrrr}
\toprule
\rot{Dataset}  & 
\rot{\begin{tabular}[c]{@{}l@{}}Invoice \\ Kickback 1\end{tabular}} & 
\rot{\begin{tabular}[c]{@{}l@{}}Invoice \\ Kickback 2\end{tabular}} & 
\rot{Larceny 1} & 
\rot{Larceny 2} & 
\rot{Larceny 3} & 
\rot{Larceny 4} & 
\rot{Larceny 5} &
\rot{Larceny 6} &
\rot{\begin{tabular}[c]{@{}l@{}}Corporate \\ Injury 1\end{tabular}} & 
\rot{\begin{tabular}[c]{@{}l@{}}Corporate \\ Injury 2\end{tabular}} & 
\rot{\begin{tabular}[c]{@{}l@{}}Selling \\ Kickback 1\end{tabular}} & 
\rot{\begin{tabular}[c]{@{}l@{}}Selling \\ Kickback 2\end{tabular}} & 
\rot{Scrap}  & 
\rot{Sale}\\
\midrule
normal 1 & 0    & 0   & 0         & 0         & 0       & 0       & 0     &   0    & 0       & 0         & 0      & 0      & 0      & 0        \\
fraud 1  & 1    & 0   & 1         & 1         & 0       & 0       & 4     &   1    & 1       & 1         & 0      & 0      & 0      & 0        \\ 
normal 2 & 0    & 0   & 0         & 0         & 0       & 0       & 0     &   0    & 0       & 0         & 0      & 0      & 3      & 2        \\
fraud 2  & 2    & 2   & 1         & 2         & 2       & 2       & 1     &   0    & 1       & 1         & 1      & 2      & 2      & 1        \\
fraud 3  & 2    & 1   & 2         & 1         & 2       & 2       & 1     &   0    & 1       & 1         & 1      & 2      & 1      & 1        \\\hline
\end{tabular}}
\end{table}


\section{Dataset Construction}
\label{chap:dataset_construction}
In the previous chapters, we discussed the generation and analysis of data from our synthetic data generation approach. 
We extract this data from the ERP system into many different tables to allow researchers to design fraud detection systems that integrate into all ERP functions as discussed in \Cref{data:requirements}.
Many fraud detection approaches, however, (especially ones utilizing machine learning \citep{schreyer2019accountingFraud}) are not capable of integrating data from multiple tables and instead require a single joint dataset to operate.
Since ERP system data is inherently multi-relational due to it tracking many different views on the underlying business activities, exactly reproducing specific joins may prove challenging. 
To allow for easy reproducibility and comparisons of ERP fraud detection approaches, we therefore focus in this section on providing single joint datasets that may be directly used for detecting fraudulent transactions.

For our joint datasets, we focus on the financial accounting data of the P2P business process in the SAP database tables RBKP, RSEG, BKPF and BSEG. We choose the financial accounting information to detect fraudulent behavior since this information is usually used by auditors in real-world auditing procedures \citep{singh2015design}.
For each run, we combine the respective tables to one dataset, obtaining a single table with financial accounting data featuring invoice, credit, general ledger (G/L) account posting and material movement transactions. 
We remove duplicate columns, empty columns, columns containing always the same value, and identifier columns, since they offer no usable information to many automatic fraud detection algorithms. 
For the remaining columns, we provide information on whether the columns contain numerical or categorical data. 

As a single fraud case may create multiple transactions within the ERP system, we mark all invoice, credit, G/L account posting and material movement transactions as fraudulent depending on whether they are part of one of our fraud scenarios from \Cref{data:frauds}, thus obtaining fraud labels for all transactions in our data.
Further, since in this join of our data we focus on financial accounting information that does not contain sales information such as final product sales prices or delivery information such as delivery addresses, we exclude the fraud cases belonging to the O2C process and the fraud cases based on delivery details since they are structurally indistinguishable from normal activity within the joint accounting tables.
The final datasets with the resulting distribution of fraudulent and total transactions are listed in \Cref{tab:data_label_dist2}.


While analysing our datasets, we found that many fraud cases are only detectable due to few anomalous entries in otherwise sparse and largely regular table data. 
This observation is also used in several well known approaches such as Benford’s law or Extreme Value Analysis, that detect anomalies by the frequency of numbers or the probability of higher values in a distribution of a single feature \citep{Barabesi2021Sep}. 
With fraud detection approaches specifically targeting few data entries of entire transactions, a fine granular labeling process that makes single fraud cases traceable on a feature level allows for gaining insights into fraud detection performance.
We therefore conduct an additional labeling process and provide additional expert feature-level annotations for each fraudulent transaction:
Within the fraudulent transactions, we identify and highlight all features that hint at the underlying fraud case due to anomalous column entries.
The resulting annotations are supplied alongside the joint datasets and may be used for in-depth prototyping and evaluation of fraud detection approaches. 
%

\begin{table}[t]
\caption{Transaction types and fraud cases of normal and partially fraudulent data.}
\label{tab:data_label_dist2}
\resizebox{\textwidth}{!}{
\begin{tabular}{lrrrrrr|rrrrrrrrr}
\toprule
\rot{Dataset}  & 
\rot{Invoice} & \rot{Credit} & 
\rot{\begin{tabular}[c]{@{}l@{}}G/L Acc.\\ Posting\end{tabular}} & 
\rot{\begin{tabular}[c]{@{}l@{}}Material \\ Receipt\end{tabular}} & 
\rot{\begin{tabular}[c]{@{}l@{}}Material \\ Withdrawl\end{tabular}} & 
\rot{\begin{tabular}[c]{@{}l@{}}Total \\ Transact.\end{tabular}} & 
\rot{\begin{tabular}[c]{@{}l@{}}Invoice \\ Kickback 1\end{tabular}} & 
\rot{\begin{tabular}[c]{@{}l@{}}Invoice \\ Kickback 2\end{tabular}} & 
\rot{Larceny 1} & \rot{Larceny 2} & \rot{Larceny 3} & \rot{Larceny 4} & 
\rot{\begin{tabular}[c]{@{}l@{}}Corporate \\ Injury\end{tabular}} & 
\rot{\begin{tabular}[c]{@{}l@{}}Frauds \\ in Total\end{tabular}}\\
\midrule
normal 1 & 7511    & 7050   & 28845               & 769              & 10495              & 54677 & 0                  & 0                  & 0         & 0         & 0         & 0         & 0                & 0               \\
fraud 1  & 5212    & 5181   & 20828               & 447              & 7758               & 39430 & 4                  & 0                  & 2         & 4         & 0         & 0         & 14               & 24              \\
normal 2 & 3231    & 3129   & 18271               & 425              & 7280               & 32337 & 0                  & 0                  & 0         & 0         & 0         & 0         & 0                & 0               \\
fraud 2  & 4154    & 4007   & 20186               & 469              & 7960               & 36778 & 6                  & 18                 & 2         & 4         & 10        & 6         & 4                & 50              \\
fraud 3  & 4080    & 3841   & 20678               & 464              & 8344               & 37407 & 24                 & 6                  & 8         & 10        & 26        & 4         & 8                & 86              \\ \hline
\end{tabular}}
\end{table}

\section{Conclusion}
\label{chap:conclusion}
With companies keeping a lock on ERP system data due to privacy and trade secrets concerns, researchers in the area of occupational fraud detection have in the past turned to synthetic data generation for validating their work.
Previous works in this area however did not provide data to the public, limiting open and reproducible research on detecting fraud in ERP systems.

In this paper, we proposed a strategy for generating ERP system data through an existing serious game, modeled a variety of occupational fraud cases within the serious game's real ERP system interface, and recorded multiple runs of both normal and fraudulent operation of a simulated make-to-stock production company.
We gave an overview of the resulting data, and provided additional joint datasets that can be directly used for applying and comparing fraud detection approaches.
Our obtained data is free from privacy and secrecy concerns and is publicly available for reproducible research on fraud detection in ERP systems.

With ERP system data of both normal and fraudulent transactions now openly available, benchmarking existing fraud detection approaches and carrying out rigorous comparisons is now a promising point for future work.
For this, we also plan on further extending the aggregation of joint datasets to enable accessible fraud detection from different perspectives of our generated ERP system data.
Further, while we provide a variety of different occupational fraud scenarios, future efforts should be directed towards the aggregation of additional novel fraud cases that allow for further validation of fraud detection performance.
Finally, since collection and annotation of datasets containing occupational fraud requires great effort, automated large scale acquisition of data for occupational fraud detection (e.g. through Active Learning) is a promising area for future work.
With this study we took a first step towards open research on ERP fraud detection and encourage future research and practical applications by providing the generated data to the general public.

\section*{Acknowledgement}
The authors acknowledge the financial support by the Federal Ministry of Education and Research of Germany as part of the DeepScan project (01IS18045A).

\printbibliography

\end{document}